\documentclass{article}
\usepackage{style/unites}
\usepackage{XCharter}
\usepackage[scaled=1.1]{zlmtt} 

\usepackage[utf8]{inputenc}
\usepackage[T1]{fontenc}
\usepackage{microtype}

\usepackage{amsmath}
\usepackage{amssymb}
\usepackage{amsfonts}
\usepackage{amsthm}
\usepackage{mathtools}
\usepackage{mathrsfs}
\usepackage{physics}
\usepackage{braket}
\usepackage{slashed}
\usepackage{nicefrac}
\usepackage{textcomp}
\usepackage{dsfont}
\usepackage{bbm}
\usepackage{bm}

\usepackage{graphicx}
\usepackage{subcaption}
\usepackage[export]{adjustbox}
\usepackage{float}
\usepackage{booktabs}
\usepackage{dcolumn}
\newcolumntype{d}[1]{D{.}{.}{#1}}
\usepackage{bigstrut, tabularx, multirow, makecell, diagbox}
\usepackage{colortbl}
\usepackage{tabularray}
\UseTblrLibrary{booktabs}
\usepackage{threeparttable}
\usepackage{tablefootnote}
\usepackage{fontawesome5}

\usepackage{placeins}
\usepackage{caption}
\usepackage{footnote}
\usepackage{enumitem}
\usepackage{multicol}
\usepackage{xspace}
\usepackage{titletoc}
\usepackage{titlesec}
\usepackage[bottom]{footmisc}
\usepackage{setspace}

\usepackage{wrapfig}
\usepackage{tikz}
\usepackage{quantikz}
\usepackage{dashbox}
\usepackage{mdframed}
\usepackage{marvosym}
\usepackage{pifont}
\usepackage{CJK}
\usepackage{url}

\usepackage[table,x11names]{xcolor}
\usepackage[most]{tcolorbox}
\tcbuselibrary{breakable}
\usetikzlibrary{decorations.pathreplacing, fit}

\definecolor{primaryblue}{HTML}{0066CC}
\definecolor{accentcyan}{HTML}{00D4AA}
\definecolor{warmorange}{HTML}{FF6B35}
\definecolor{deepgray}{HTML}{2C3E50}
\definecolor{lightgray}{HTML}{F8F9FA}
\definecolor{gradientstart}{HTML}{667eea}
\definecolor{gradientend}{HTML}{764ba2}

\definecolor{citecolor}{HTML}{0071bc}
\definecolor{citeblue}{RGB}{0, 113, 188}
\definecolor{linkcolor}{HTML}{9A4D92}
\definecolor{firebrick}{rgb}{0.698,0.133,0.133}

\definecolor{paleviolet}{HTML}{E1EEFC}
\definecolor{CarolinaUltraLight}{HTML}{E7F4FC}
\definecolor{lightgrey}{RGB}{247, 247, 247}
\definecolor{shadecolor}{HTML}{EFEFEF}
\definecolor{lightyellow}{rgb}{1.0, 0.95, 0.7}
\definecolor{lightblue}{rgb}{0.90, 0.95, 1.0}
\definecolor{light-gray}{gray}{0.95}

\definecolor{darkgrey}{rgb}{0.5, 0.5, 0.5}
\definecolor{darkgreen}{rgb}{0, 0.5, 0}
\definecolor{mydarkblue}{rgb}{0,0.08,0.45}
\definecolor{mydarkblue2}{rgb}{0.133, 0.133, 0.698}
\definecolor{echodrk}{HTML}{0099cc}
\definecolor{mymauve}{rgb}{0.58,0,0.82}
\definecolor{midnightblue}{rgb}{0.1,0.1,0.44}
\definecolor{oxfordblue}{rgb}{0.0,0.13,0.28}
\definecolor{prussianblue}{rgb}{0.0,0.19,0.33}
\definecolor{coolteal}{rgb}{0, 0.45, 0.45}
\definecolor{olive}{rgb}{0.1, 0.3, 0}
\definecolor{mypurple}{rgb}{0.5,0,0.5}
\definecolor{almond}{rgb}{0.94, 0.87, 0.8}

\definecolor{blue_ampEncoding}{HTML}{DAE8FC}
\definecolor{green_encoder}{HTML}{D5E8D4}
\definecolor{purple_decoder}{HTML}{E1D5E7}
\definecolor{yellow_measure}{HTML}{FFF2CC}
\definecolor{gray_block}{HTML}{F5F5F5}
\definecolor{pink_dru}{HTML}{FAD9D5}
\definecolor{orange_v}{HTML}{FAD7AC}

\definecolor{colorA}{rgb}{1,0,0}
\definecolor{colorB}{rgb}{0,0.3,1}
\definecolor{colorC}{rgb}{0.9,0.8,0.2}
\definecolor{colorD}{rgb}{0,0.65,0}
\definecolor{lesslightgray}{rgb}{0.5,0.5,0.5}
\definecolor{fundamental}{RGB}{55, 110, 111}
\definecolor{Gred}{RGB}{219, 50, 54}
\definecolor{ToCgreen}{RGB}{0, 128, 0}
\definecolor{Sepia}{RGB}{112, 66, 20}
\definecolor{Dblue}{rgb}{0,0.08,0.45}
\definecolor{Blue}{rgb}{0, 0, 0.8}
\definecolor{blue}{rgb}{0,0,1}
\definecolor{UNCblue!10}{rgb}{0.84,0.91,0.98}
\definecolor{RowAlt}{rgb}{0.98,0.98,0.99}

\definecolor{CarolinaBlue}{HTML}{7BAFD4}        
\definecolor{CarolinaLightBlue}{HTML}{B3D4E5}   
\definecolor{CarolinaUltraLight}{HTML}{E8F4F8}  
\definecolor{CarolinaText}{HTML}{1C2B33}        

\usepackage[pagebackref=true,breaklinks=true,colorlinks,hyperfootnotes=false]{hyperref}
\hypersetup{
  colorlinks,
  citecolor=citeblue,
  linkcolor=firebrick,
  urlcolor=firebrick
}
\usepackage[nameinlink,capitalize,noabbrev]{cleveref}

\titlespacing\section{0pt}{4pt plus 4pt minus 2pt}{-2pt plus 2pt minus 2pt}
\titlespacing\subsection{0pt}{2pt plus 4pt minus 2pt}{-2pt plus 2pt minus 2pt}
\titlespacing\subsubsection{0pt}{2pt plus 4pt minus 2pt}{-2pt plus 2pt minus 2pt}

\makeatletter
\def\th@remark{%
  \thm@headfont{\bfseries}%
  \normalfont 
  \thm@preskip\topsep \divide\thm@preskip\tw@
  \thm@postskip\thm@preskip
}
\makeatother

\theoremstyle{definition}

\tcolorboxenvironment{theorem}{
  breakable,
  colback=black!10,
  colframe=white,
  width=\linewidth, 
  enlarge left by=0pt,
  enlarge right by=0pt,
  boxsep=5pt,
  boxrule=0pt,
  left=0pt,right=0pt,top=0pt,bottom=0pt,
  arc=8pt,
  before skip=\topsep,
  after skip=\topsep
}

\tcolorboxenvironment{lemma}{
  breakable,
  colback=black!10,
  colframe=white,
  width=\linewidth,
  enlarge left by=0pt,
  enlarge right by=0pt,
  boxsep=5pt,
  boxrule=0pt,
  left=0pt,right=0pt,top=0pt,bottom=0pt,
  arc=8pt,
  before skip=\topsep,
  after skip=\topsep
}

\tcolorboxenvironment{corollary}{
  breakable,
  colback=black!10,
  colframe=white,
  width=\linewidth,
  enlarge left by=0pt,
  enlarge right by=0pt,
  boxsep=5pt,
  boxrule=0pt,
  left=0pt,right=0pt,top=0pt,bottom=0pt,
  arc=8pt,
  before skip=\topsep,
  after skip=\topsep
}

\tcolorboxenvironment{proposition}{
  breakable,
  colback=black!10,
  colframe=white,
  width=\linewidth,
  enlarge left by=0pt,
  enlarge right by=0pt,
  boxsep=5pt,
  boxrule=0pt,
  left=0pt,right=0pt,top=0pt,bottom=0pt,
  arc=8pt,
  before skip=\topsep,
  after skip=\topsep
}

\newtheorem{definition}{Definition}[section]
\tcolorboxenvironment{definition}{
  breakable,
  colback=black!10,
  colframe=white,
  width=\linewidth,
  enlarge left by=0pt,
  enlarge right by=0pt,
  boxsep=5pt,
  boxrule=0pt,
  left=0pt,right=0pt,top=0pt,bottom=0pt,
  arc=8pt,
  before skip=\topsep,
  after skip=\topsep
}

\tcolorboxenvironment{assumption}{
  breakable,
  colback=black!10,
  colframe=white,
  width=\linewidth,
  enlarge left by=0pt,
  enlarge right by=0pt,
  boxsep=5pt,
  boxrule=0pt,
  left=0pt,right=0pt,top=0pt,bottom=0pt,
  arc=8pt,
  before skip=\topsep,
  after skip=\topsep
}

\tcolorboxenvironment{claim}{
  breakable,
  colback=black!10,
  colframe=white,
  width=\linewidth,
  enlarge left by=0pt,
  enlarge right by=0pt,
  boxsep=5pt,
  boxrule=0pt,
  left=0pt,right=0pt,top=0pt,bottom=0pt,
  arc=8pt,
  before skip=\topsep,
  after skip=\topsep
}

\tcolorboxenvironment{problem}{
  breakable,
  colback=black!10,
  colframe=white,
  width=\linewidth,
  enlarge left by=0pt,
  enlarge right by=0pt,
  boxsep=5pt,
  boxrule=0pt,
  left=0pt,right=0pt,top=0pt,bottom=0pt,
  arc=8pt,
  before skip=\topsep,
  after skip=\topsep
}

\tcolorboxenvironment{question}{
  breakable,
  colback=black!10,
  colframe=white,
  width=\linewidth,
  enlarge left by=0pt,
  enlarge right by=0pt,
  boxsep=5pt,
  boxrule=0pt,
  left=0pt,right=0pt,top=0pt,bottom=0pt,
  arc=8pt,
  before skip=\topsep,
  after skip=\topsep
}

\newtcolorbox{titleblock}{
  enhanced,
  frame hidden,
  colback=CarolinaUltraLight,
  colframe=CarolinaUltraLight,
  boxrule=0pt,
  arc=10pt,
  left=14pt,
  right=14pt,
  top=14pt,
  bottom=14pt,
  width=\linewidth,
  before skip=12pt plus 4pt,
  after skip=12pt plus 4pt,
  grow to left by=1.5pt,
  grow to right by=1.5pt,
  before upper={
    \setlength{\parindent}{0cm}
    \setlength{\parskip}{0.5cm}
  }
}

\crefname{theorem}{Theorem}{Theorems}
\crefname{proposition}{Proposition}{Propositions}
\crefname{lemma}{Lemma}{Lemmas}
\crefname{corollary}{Corollary}{Corollaries}
\crefname{definition}{Definition}{Definitions}
\crefname{assumption}{Assumption}{Assumptions}
\crefname{remark}{Remark}{Remarks}
\crefname{problem}{Problem}{Problems}
\crefname{property}{Property}{property}
\crefname{question}{Question}{Questions}

\numberwithin{equation}{section}
\numberwithin{theorem}{section}
\numberwithin{proposition}{section}
\numberwithin{definition}{section}
\numberwithin{lemma}{section}
\numberwithin{assumption}{section}
\numberwithin{remark}{section}

\newcommand\metadataformat[2][]{{\small {\bfseries #1:} #2}}

\def\1{\bm{1}}

\makeatletter
\let\save@mathaccent\mathaccent
\newcommand*\if@single[3]{%
    \setbox0\hbox{${\mathaccent"0362{#1}}^H$}%
    \setbox2\hbox{${\mathaccent"0362{\kern0pt#1}}^H$}%
    \ifdim\ht0=\ht2 #3\else #2\fi
}
\newcommand*\rel@kern[1]{\kern#1\dimexpr\macc@kerna}
\newcommand*\widebar[1]{\@ifnextchar^{{\wide@bar{#1}{0}}}{\wide@bar{#1}{1}}}
\newcommand*\wide@bar[2]{\if@single{#1}{\wide@bar@{#1}{#2}{1}}{\wide@bar@{#1}{#2}{2}}}
\newcommand*\wide@bar@[3]{%
    \begingroup
    \def\mathaccent##1##2{%
        \let\mathaccent\save@mathaccent
        \if#32 \let\macc@nucleus\first@char \fi
        \setbox\z@\hbox{$\macc@style{\macc@nucleus}_{}$}%
        \setbox\tw@\hbox{$\macc@style{\macc@nucleus}{}_{}$}%
        \dimen@\wd\tw@
        \advance\dimen@-\wd\z@
        \divide\dimen@ 3
        \@tempdima\wd\tw@
        \advance\@tempdima-\scriptspace
        \divide\@tempdima 10
        \advance\dimen@-\@tempdima
        \ifdim\dimen@>\z@ \dimen@0pt\fi
        \rel@kern{0.6}\kern-\dimen@
        \if#31
        \overline{\rel@kern{-0.6}\kern\dimen@\macc@nucleus\rel@kern{0.4}\kern\dimen@}%
        \advance\dimen@0.4\dimexpr\macc@kerna
        \let\final@kern#2%
        \ifdim\dimen@<\z@ \let\final@kern1\fi
        \if\final@kern1 \kern-\dimen@\fi
        \else
        \overline{\rel@kern{-0.6}\kern\dimen@#1}%
        \fi
    }%
    \macc@depth\@ne
    \let\math@bgroup\@empty \let\math@egroup\macc@set@skewchar
    \mathsurround\z@ \frozen@everymath{\mathgroup\macc@group\relax}%
    \macc@set@skewchar\relax
    \let\mathaccentV\macc@nested@a
    \if#31
    \macc@nested@a\relax111{#1}%
    \else
    \def\gobble@till@marker##1\endmarker{}%
    \futurelet\first@char\gobble@till@marker#1\endmarker
    \ifcat\noexpand\first@char A\else
    \def\first@char{}%
    \fi
    \macc@nested@a\relax111{\first@char}%
    \fi
    \endgroup
    }
\makeatother
\let\bar\widebar

\DeclareMathAlphabet{\mathsfit}{\encodingdefault}{\sfdefault}{m}{sl}
\SetMathAlphabet{\mathsfit}{bold}{\encodingdefault}{\sfdefault}{bx}{n}

\let\tilde\widetilde
\let\hat\widehat

\renewcommand{\arraystretch}{1.15}
\usepackage{listings}
\newcommand{\ours}{\texttt{SkillComposer}\xspace}

\begin{document}

\makeatletter
\def\blfootnote{\gdef\@thefnmark{}\@footnotetext}
\makeatother

\makeatletter
\pagestyle{fancy}
\fancyhf{}
\renewcommand{\headrulewidth}{1pt}
\chead{\small\bf Generative Skill Composition for LLM Agents
}
\cfoot{\thepage}
\thispagestyle{fancy}
\makeatother

\makeatletter
\def\icmldate#1{\gdef\@icmldate{#1}}
\icmldate{\today}
\makeatother

\makeatletter
\fancypagestyle{fancytitlepage}{
  \fancyhead{}
  \lhead{\includegraphics[height=0.8cm]{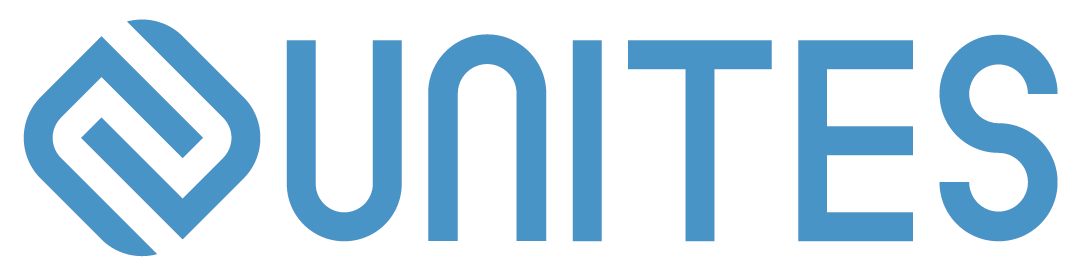}}
  \rhead{\it \@icmldate}
  \cfoot{}
}
\makeatother

\thispagestyle{fancytitlepage}

\vspace*{0.5em}

\noindent
\begin{titleblock}
    {\setlength{\parskip}{0cm}
     \raggedright
     {\setstretch{1.2}
      \LARGE\bfseries
      Generative Skill Composition for LLM Agents
      \par}
    }
    \vskip 0.2cm
    
    \begin{icmlauthorlist}
\mbox{Xinyu Zhao$^{\,1}$},
\mbox{Zhen Tan$^{\,2}$},
\mbox{Vaishnav Tadiparthi$^{\,3}$},
\mbox{Nakul Agarwal$^{\,3}$},
\mbox{Kwonjoon Lee$^{\,3}$},
\mbox{Ehsan Moradi Pari$^{\,3}$},
\mbox{Hossein Nourkhiz Mahjoub$^{\,3}$}
and \mbox{Tianlong Chen$^{\,1\,\textrm{\Letter}}$}
\end{icmlauthorlist}

$^{1\,}$University of North Carolina at Chapel Hill \quad $^{2\,}$Arizona State University \quad $^{3\,}$Honda Research Institute USA


    \vskip 0.2cm

{\noindent\bfseries\large Abstract\par}
Recent LLM agents benefit from skills for solving complex tasks. Skills encapsulate modular packages of procedural knowledge and instructions for performing specialized tasks, such as setting up a sandboxed environment, running a test suite, or refactoring a function across multiple files. As skill libraries grow and become reusable across tasks and domains, selecting an appropriate skill composition has emerged as a central bottleneck. Existing approaches fall into two categories. One exposes the agent's reasoning to the entire skill collection; the other performs skill retrieval via embeddings or LLM-based rerankers. Both provide useful insights; however, they miss the structural nature of skill composition, which is a joint decision over which skills, how many, and in what order---three dimensions that cannot be decoupled. We formalize this as structured skill composition: given a task and a skill library, predict an executable skill plan that jointly specifies the activated subset, count, and execution order. We propose \ours, which instantiates structured skill composition as task-conditioned skill sequence prediction. \ours uses a constrained autoregressive decoder over skill identifiers, so subset, count, and order emerge jointly from a single decoding pass, and dependencies between successive skills are captured naturally. We build a training set of task--composition pairs from a real, human-curated skill library. We then evaluate \ours along two axes: composition quality on a held-out test set, and downstream task success on SkillsBench across two production-grade coding agents. On \{GPT-5.2-Codex, Gemini-3-Pro-Preview\}, \ours raises the pass rate by \{$+23.1$, $+18.2$\}\,pp over the no-skill baseline, surpassing top-3 retrieval and matching the gold-skill retrieval upper bound at lower prompt-token cost.

    \vskip 0.2cm
    {\setlength{\parskip}{0cm}
     \centering
     \makebox[\linewidth]{
        \metadataformat[Project Page]{
            \href{https://skill-composer.github.io/}{https://skill-composer.github.io/}
        }
     }
    }
\end{titleblock}

\blfootnote{%
$^{\textrm{\Letter}}$ Corresponding authors
\\[2.5em]
\ifcsname @icmlpreprint\endcsname
  \textit{\csname @icmlpreprint\endcsname}%
\fi
}

\section{Introduction}
\label{sec:intro}

Skill libraries allow LLM agents to apply procedural knowledge across complex tasks such as structured document generation, software development, and computer-use automation~\cite{jimenez2024swebench,yang2024sweagent,xie2024osworld,zhou2024webarena,trivedi2024appworld}.
A skill is a modular, reusable unit of procedural knowledge, \textit{e.g.} bundling natural-language instructions, scripts, and supporting resources that an agent dynamically loads into its working context to perform a specialized subtask, such as coordinating code review and test execution, generating artifacts, or producing structured documents~\cite{anthropic2025skills,xu2026agentskillsurvey,jiang2026sok}.
The research community has made substantial progress on curating skills for varied tasks~\cite{wang2023voyager,chen2026skillx,xia2026skillrl,li2025skillflow,jiao2026agentproposer,liu2025toollibgen,liu2026graphofskills}, driving skill libraries to grow rapidly in size.
This shifts the inference-time bottleneck from obtaining skills to composing the right collection of skills: to improve task performance, the agent must decide which skills to load, how many, and in what order they should be used.

Current approaches for selective skill use fall into two paradigms.
Retrieval ranks skills independently using LLM-as-a-judge or task-skill embedding similarity~\cite{zheng2026skillrouter,su2026skillretrieval,li2025skillflow}, while end-to-end planning exposes the agent to the full skill library, where the agent needs to reason to trigger appropriate skills at the same time as approaching input tasks~\cite{wang2023voyager,yao2023react,schick2023toolformer,xia2026skillrl}.
These interfaces are sufficient when a task maps cleanly to a single dominant skill, but they leave important structural aspects implicit in compositional tasks.
For example, consider a request to ``locate a deprecated API call, refactor it across the codebase, and run the regression suite.''
A useful plan should first identify the relevant call sites, then apply the refactor, and finally run tests to validate the change.
Retrieval can surface individually relevant skills for search, editing, and testing, but a ranked list alone does not specify how many skills should be used or the order in which they should be executed.

\begin{figure}[t]
    \centering
    \includegraphics[width=\linewidth]{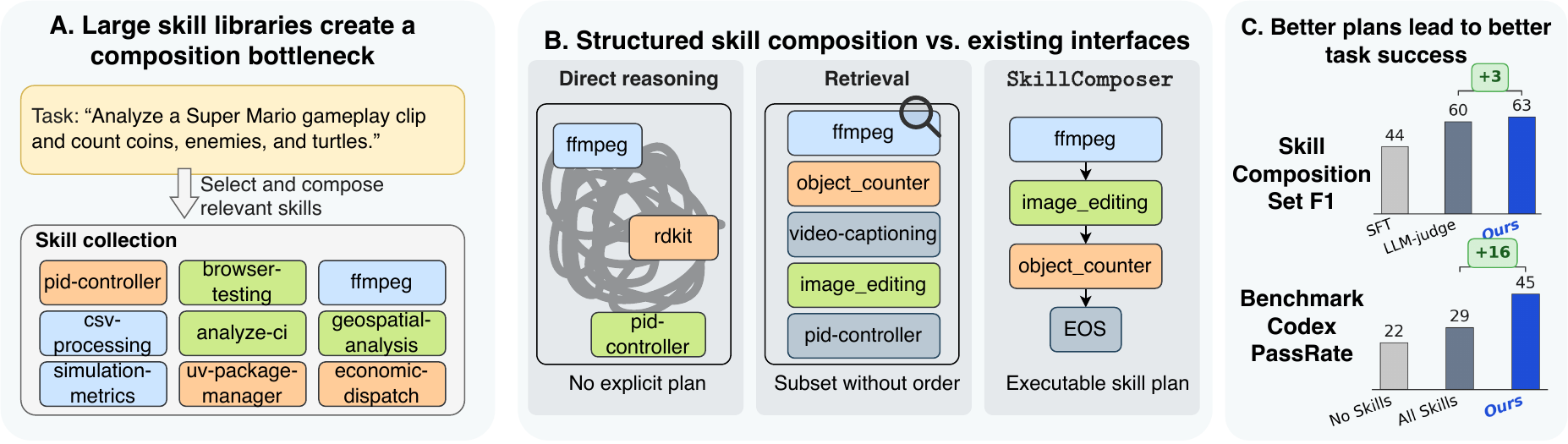}
    \caption{{\small \textbf{Structured skill composition with \ours.}
    \textbf{(A)} Large skill libraries create a composition bottleneck. To solve complex tasks, an agent must decide not only which skills to use, but also their exact count and execution order.
    \textbf{(B)} Existing paradigms: directly exposing the agent to all skill options leaves composition implicit within an unstructured execution trace, while retrieval methods only return an unordered subset of candidates. We propose \ours, which explicitly predicts an ordered, executable skill sequence.
    \textbf{(C)} By structuring the composition process, \ours  improves both plan exact match and downstream task success rates on SkillsBench.}}
    \vspace{-4mm}
    \label{fig:teaser}
\end{figure}

We formulate structured skill composition as task-conditioned skill sequence prediction.
Given a natural-language task and a fixed reusable skill library, the model predicts an ordered sequence of skill identifiers ending with a stop symbol.
We propose \ours, a generative framework that uses a constrained autoregressive decoder over existing library skills, where subset selection, skill-count prediction, and skill ordering emerge jointly from a single decoding pass.
Because each prediction is conditioned on the task, the skill library, and the previously selected skills, \ours can capture dependencies between successive skills without requiring an explicit execution order at inference time.
Also, the output vocabulary is a valid skill identifier, the predicted plan is inspectable, and can be loaded directly into downstream agents.

To build \ours, we construct a dataset of task-skill composition pairs grounded in a real, human-curated skill library.
Starting from real task-skill composition seeds from SkillBench~\cite{li2026skillsbench}, we build a skill dependency graph using skill metadata and observed workflow co-occurrence.
We then use layered synthesis and filtering to obtain supervision for both single-skill grounding and multi-skill compositions.
This construction is designed to cover long-tail individual skills and dependency-aware skill chains, while keeping the output space tied to executable skills in the library.

We evaluate \ours along two complementary axes.
First, we measure composition quality on evaluation sets from both synthetic and held-out real data, testing whether the predicted plan matches the target skill subset, count, and order.
Second, we measure downstream task success on SkillsBench across two production-grade coding agents (GPT-5.2-Codex, Gemini-3-Pro-Preview) to test whether better skill plans translate into better agent execution.
Our main contributions are:

\begin{itemize}[leftmargin=1em]
    \item [$\star$] We formalize inference-time skill use for LLM agents as a structured prediction problem over a fixed skill library, where the output plan jointly determines which skills to activate, how many skills are needed, and in what order they should be executed.
    \vspace{-1.25mm}

    \item [$\star$] We construct a dataset grounded in a real, human-curated skill library. Starting from real task-composition seeds, we build a skill dependency graph and use layered synthesis with quality filtering to obtain supervision for both single-skill and multi-skill dependency-aware composition.
    \vspace{-1.25mm}

    \item [$\star$] We propose \ours, a task-conditioned skill sequence predictor with a constrained autoregressive decoder over skill identifiers. \ours unifies subset selection, cardinality prediction, and ordering into a single decoding process while ensuring that every generated element corresponds to an executable library skill.
    \vspace{-1.25mm}

    \item [$\star$] We evaluate \ours on composition quality and downstream task success across two production-grade coding agents. On \{Codex, Gemini\}, \ours raises SkillsBench pass rate by \{$+23.1$, $+18.2$\}\,pp over the no-skill baseline, surpassing retrieval and matching the gold-skill retrieval upper bound at lower prompt-token cost.
    \vspace{-1.5mm}
\end{itemize}

\section{Related Works}
\label{sec:relwork}

\paragraph{Skill libraries and discovery.}
A growing line of work equips LLM agents with reusable skills or tools that can be retrieved at deployment time~\cite{wang2023voyager,yuan2023craft,qian2023creator,ma2025automated}.
While these methods have shown clear benefits of maintaining a skill inventory, they all adopt flat retrieval at selection time, ranking skills independently by embedding similarity without considering how many skills are needed or how they depend on one another.
In contrast, our work takes a curated skill library as given and addresses the complementary problem of structured composition: jointly predicting the skill subset, the number of skills, and execution order. Recent work further validates the growing importance of skill-based agent design across retrieve-and-rerank routing, retrieval-augmented skill use, benchmarking, lifecycle taxonomies, RL-based skill construction, and analyses under realistic retrieval conditions~\cite{zheng2026skillrouter,li2025skillflow,liu2026graphofskills,su2026skillretrieval,li2026skillsbench,jiang2026sok,xia2026skillrl,liu2026agent}.
None of these efforts model skill selection as closed-vocabulary sequence generation with explicit cardinality and ordering decisions.

\vspace{-2mm}

\paragraph{Tool-level planning and composition.}
A parallel research thread plans over tool or API action spaces at the atomic function-call level, casting selection and ordering as decomposition, search, graph evaluation, vocabulary embedding, or graph-structured planning~\cite{shen2023hugginggpt,zhuang2023toolchain,shen2024taskbench,hao2023toolkengpt,wu2024can}.
Although these systems also reason about inter-step dependencies, they operate at the API-call level, where typed signatures and return values provide strong structural signals for selection and ordering.
Our work operates at a coarser skill level, where each skill encapsulates a reusable multi-step procedure spanning several API calls (e.g., ``schema grounding'' or ``query reformulation''), and this shift introduces challenges API-level methods do not face: skills lack typed signatures, so dependencies are latent and task-logical rather than type-induced, and the catalog is small but highly interacting, swapping or reordering even two skills can flip task outcome.
Flat similarity retrieval and atomic-action search are thus both ill-suited, motivating skill composition as joint prediction.

\section{Preliminaries}
\label{sec:preliminaries}
\vspace{-1mm}

Following the open Agent Skills standard~\footnote{https://agentskills.io/what-are-skills} and \cite{jiang2026sok}, an agentic skill is a reusable procedural module that augments agent behavior at inference time by injecting procedural knowledge into the prompt context, without modifying model weights. Formally:

\begin{definition}[\textbf{Skills}]
We define each skill as $s_i=(m_i, C_i, \pi_i, T_i, R_i)$,
where $m_i$ is metadata (name + one-line description, e.g.\ ``\textit{flood-detection -- compare water levels to thresholds; count flood days}), $C_i$ is an applicability condition (e.g.``\textit{input contains a time-indexed water-level series and station thresholds}''), $\pi_i$ is a procedural policy (e.g.``\textit{aggregate instantaneous values to daily extremes, then compare against the action/minor/moderate/major bands}''), $T_i$ is a termination condition (e.g., ``\textit{a per-station flood\_days count has been written to the requested output path}'')), and $R_i$ is an optional callable interface or supporting resource (e.g.``\textit{a Python helper, a REST endpoint such as the USGS dataretrieval API, or a bundled lookup table}'').
\end{definition}
Unlike tools (atomic API calls), plans (ephemeral task-specific reasoning), and prompt templates (static text without applicability gating), a skill encodes reusable procedural knowledge that persists across tasks and sessions.
We further define a skill library as follows:
\begin{definition}[\textbf{Skill Library}]
{Let $\mathcal{S}=(s_1,\dots,s_K)$ denote a library of $K$ reusable skills.
Each skill $s_i$ exposes metadata $m_i$ for runtime discovery via progressive disclosure: agents load compact metadata at startup and activate full instructions on demand.}
\end{definition}
We treat $\mathcal{S}$ as fixed at training time because real agentic deployments ship with curated skill packs rather than open-ended skill generation~\cite{li2026skillsbench}, thereby isolating the composition problem from the orthogonal skill-creation problem.
With these, we can formulate our problem:
\begin{definition}[\textbf{Task-conditioned Skill Sequence Prediction}]

Given a task description \(x \in \mathcal{X}\), an environment context \(c \in \mathcal{C}\), and a skill library \(\mathcal{S}=\{s_1,\ldots,s_K\}\), we formulate skill composition as task-conditioned skill sequence prediction. A parameterized model \(f_\theta\) predicts a variable-length sequence of skill indices:
\begin{equation}
\hat{\mathbf{z}} = (\hat{z}_1, \hat{z}_2, \ldots, \hat{z}_{\hat{n}}, \texttt{STOP})
= f_\theta(x, c, \mathcal{S})
\label{eq:skill-sequence-prediction}
\end{equation}
where \(\theta\) are learnable parameters, each \(\hat{z}_t \in \{1,\dots,K\}\) indexes a skill in \(\mathcal{S}\), and \(\texttt{STOP}\) is a special end-of-sequence symbol that signals termination. The predicted skill count \(\hat{n}\) is determined by the position of \(\texttt{STOP}\). The predicted skill sequence is then recovered by index lookup:
$
\hat{\mathbf{s}} = (s_{\hat{z}_1}, s_{\hat{z}_2}, \dots, s_{\hat{z}_{\hat{n}}}).
$
\end{definition}
The prediction jointly resolves three coupled aspects: \textit{(1) subset selection} — which skills are relevant; \textit{(2) skill count} — how many skills are needed (task-dependent, not fixed or predefined); and \textit{(3) ordering} — in what sequence the selected skills should be composed.
At inference time, the resolved skills $\hat{\mathbf{s}}$ are loaded into the agent's context in the predicted order, providing procedural guidance for downstream execution.
Figure~\ref{fig:prelim_example} illustrates a concrete instance: the model receives a task with the skill library (each entry as a compact \textit{name + description}), and outputs a short index sequence ending in \texttt{STOP}, which is resolved to full skill instructions before being prepended to the agent's prompt.

\begin{figure}[t]
\centering
\fbox{%
\parbox{0.95\linewidth}{\footnotesize
\textbf{Input:} task $x$, environment $c$, skill library
$\mathcal{S}=\{s_1,\dots,s_K\}$ with $K = 196$.\\[2pt]

\texttt{[Skill library -- name + one-line description per entry]}\\
\begin{tabular}{@{\quad}l@{\;\;}l}
$s_{55}$  & \texttt{flood-detection} -- compare water levels to thresholds; count flood days.\\
$s_{104}$ & \texttt{nws-flood-thresholds} -- fetch action/minor/moderate/major flood stages from NWS.\\
$s_{184}$ & \texttt{usgs-data-download} -- pull USGS streamflow / gage-height series.\\
$\dots$   & 193 other skills, e.g.\ \texttt{pdf}, \texttt{rdkit}, \texttt{pptx}, $\dots$\\
\end{tabular}\\[2pt]

\texttt{[Task $x$]} ``Find Michigan USGS stations that experienced flooding during April 1--7, 2025; write
(\texttt{station\_id}, \texttt{flood\_days}) to \texttt{flood\_results.csv}.''\\

\texttt{[Environment $c$]} \texttt{/root/data/michigan\_stations.txt} (streamflow records).\\[3pt]

\textbf{Predicted sequence:}
$(s_{104},\, s_{184},\, s_{55}) \to
(\texttt{nws-flood-thresholds}, \texttt{usgs-data-download}, \texttt{flood-detection})$

}%
}
\caption{\small Example of selecting an ordered skill sequence from a large skill library given a task and environment.}
\label{fig:prelim_example}
\vspace{-3mm}
\end{figure}

\section{\ours: Generative Composition of Skill Sequences}
\label{sec:method}

\subsection{Overview}
\label{sec:method_overview}

Inspired by recent work that frames retrieval and tool selection as generation over a closed vocabulary~\cite{rajput2023recommender,hao2023toolkengpt}, we treat each library index $i\in\{1,\dots, K\}$ as a primitive output token of a small task-conditioned decoder, with \texttt{STOP} marking sequence termination.
This single sequence model jointly resolves the three coupled aspects identified in \S\ref{sec:preliminaries}: \textit{which} skills (subset selection), \textit{how many} (skill count), and \textit{in what order} (procedural ordering).
Taking the flood-analysis task in Figure~\ref{fig:prelim_example} as an example, the skill composing model reads the task $x$, the environment $c$ (the Michigan stations file), and the compact metadata $\{m_i\}_{i=1}^{K}$, and emits the index sequence $(104,\,184,\,55,\,\texttt{STOP})$.
Index lookup recovers the ordered plan $\hat{\mathbf{s}} = (s_{104},\,s_{184},\,s_{55}) = (\texttt{nws-flood-thresholds},\,\texttt{usgs-data-download},\,\texttt{flood-detection})$.

\begin{figure}[t]
    \centering
    \includegraphics[width=0.9\linewidth]{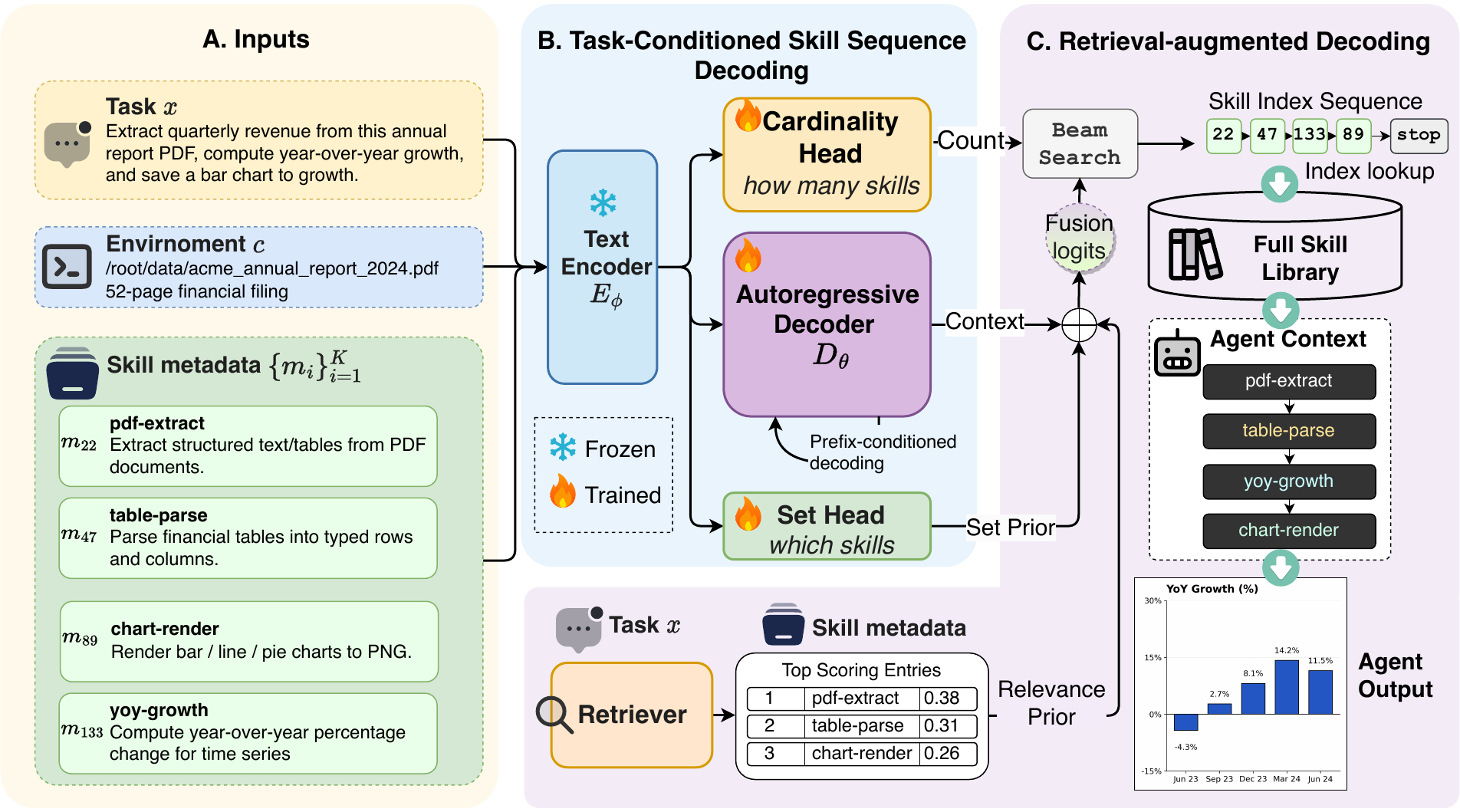}
    \caption{{\small
    \ours method overview. \textbf{(A)} Given a task, environment context, and compact metadata from a fixed skill library, \ours encodes the task–library context and predicts a variable-length ordered sequence of skill indexes. \textbf{(B)} The autoregressive decoder produces contextual skill logits, while auxiliary cardinality and set heads estimate how many skills are needed and which skills are relevant. \textbf{(C)} During retrieval-augmented decoding, contextual logits are fused with retrieval and set-membership priors, and the predicted count guides length / STOP control. The resulting skill-index sequence is resolved into full skill packages and loaded into the downstream agent context in the predicted order.}}
    \label{fig:method}
    \vspace{-4mm}
\end{figure}

\ours has three components, illustrated in Figure~\ref{fig:method}: (\textit{i}) a frozen text encoder that maps the serialized prompt $P(x,c,\mathcal{S})$ into a dense task representation; (\textit{ii}) a constrained autoregressive decoder with auxiliary cardinality and set-membership heads; and (\textit{iii}) a retrieval-augmented decoding step that fuses a lexical similarity prior into the skill logits.
We describe each in turn.

\subsection{Task-Conditioned Composer}
\label{sec:method_composer}
\paragraph{Skill tokenization.} We start with tokenizing each library skill $s_i$ by its integer index $i\in\{1,\dots,K\}$, yielding the output vocabulary $\mathcal{V}=\{1,\dots,K\}\cup\{\texttt{STOP}\}$.
Only the compact metadata $m_i$ is consumed during prediction; the full procedural policy $\pi_i$ and resources $R_i$ are activated only after $\hat{\mathbf{s}}$ is resolved, mirroring progressive disclosure in skill libraries.

\paragraph{Task Encoder.}
A frozen pretrained text encoder $E_\phi$ maps the serialized prompt $P(x,c,\mathcal{S})$ into a pooled task vector $\mathbf h = W_{\text{proj}}\mathbf h_x\in\mathbb R^d$.
We instantiate $E_\phi$ as Qwen3-Embedding-0.6B (last-token pooled, output dimension 1024) and project to $d=256$.
The encoder parameters $\phi$ are frozen. The projection $W_{\text{proj}}$, decoder parameters $\theta$, and auxiliary-head parameters $\psi,\xi$ are trained.

\paragraph{Autoregressive decoder.}
A small transformer decoder $D_\theta$ predicts each token conditioned on the projected task vector and all previously generated tokens $\hat{z}_{<t} := (\hat{z}_1,\dots,\hat{z}_{t-1})$:
\begin{equation}
p_\theta(\mathbf z\mid x,c,\mathcal S)
=
\prod_{t=1}^{n+1}
p_\theta
\left(
z_t
\mid
\mathbf h,\mathbf z_{<t}
\right)
\label{eq:ar}
\end{equation}
We instantiate $D_\theta$ as a 3-layer, 256-dim transformer with 4 attention heads.
Skills $s_i$ are surfaced to the decoder via cross-attention to projected metadata embeddings, so that the decoder distinguishes the $K$ candidates by their natural-language descriptions rather than their indexes alone.

\paragraph{Factorized supervision via auxiliary heads.}
The autoregressive head models the joint distribution over the three aspects of skill composition identified in \S\ref{sec:method_overview}: \textit{which} skills, \textit{how many}, and \textit{in what order}.
This joint formulation is desirable for ordering, where conditioning on $\hat{z}_{<t}$ is essential, but it dilutes the supervision available to the other two aspects: the entire length signal is carried by a single \texttt{STOP} position, and the positive supervision for a relevant skill appears only at its gold position. The model receives no order-agnostic signal that the skill is relevant, independent of where it appears.
We therefore complement the AR head with two task-conditional auxiliary heads, one per remaining aspect, each attached to the task vector $\mathbf{h}_x$ and trained jointly with the sequence loss.
The AR head retains responsibility for ordering, while cardinality and set membership receive dedicated supervision channels and can also be reused as decoding priors at inference time.

\textit{Cardinality head (how many).}
A linear classifier on $\mathbf{h}_x$ predicts the skill count $\hat{n}\in\{1,\dots,N_\text{max}\}$ directly (we set $N_\text{max}{=}8$),
\begin{equation}
p_\psi(\hat{n}\mid x,c) \;=\; \texttt{softmax}(W_n\,\mathbf{h})
\label{eq:card_head}
\end{equation}
yielding a length signal that is independent of the AR head's emission of \texttt{STOP} and can be used to softly bias or hard-clip decoding.

\textit{Set head (which skills).}
A pairwise matcher $g_\xi$ scores each library skill $s_i$ independently against the task vector,

\begin{equation}
\sigma_i=g_\xi(\mathbf h,\mathbf e_i)
=
\texttt{MLP}_{\xi}\!\left(
[\mathbf h;\mathbf e_i;\mathbf h\odot\mathbf e_i;|\mathbf h-\mathbf e_i|]
\right)
\label{eq:set_head}
\end{equation}
where $\mathbf e_i=W_m E_\phi(m_i)\in\mathbb R^d$ is the projected metadata embedding of $s_i$ and the four concatenated terms capture identity, interaction, and distance between task and skill.
$\texttt{MLP}_{\xi}$ is a 2-layer MLP with hidden width 256 and a single output logit.
Supervision is binary cross-entropy against the gold membership indicator
$\mathbbm{1}[s_i\in\hat{\mathbf{s}}]$
, so every relevant skill receives a direct gradient independent of its position in $\mathbf{z}$.

\subsection{Retrieval-Augmented Decoding}

The autoregressive decoder produces a contextual prediction that conditions every step on the previously decoded output, routing all task information through the dense vector $\mathbf{h}_x$ and learned skill embeddings. Beyond this contextual channel, two structural properties of skills motivate an additional inference-time prior.
First, the skill library is heavy-tailed: many skills appear in only one or two training tasks (\S\ref{sec:data}). Learned representations have a weak signal for discriminating them, whereas retrieval scores draw on the full library corpus and require no per-skill training data.
Second, the skill vocabulary structure makes such an inference-time prior essentially free: each output index corresponds to a fixed metadata document, so any task-skill relevance scorer can be precomputed once per task and reused across all decoding steps without modifying the decoder.
We therefore complement the contextual channel with a second, position-independent channel that scores the standalone semantic relevance of each library skill to the task: how well skill $s_i$ matches $(x,c)$ on its own merits.
Together, the decoder captures contextual ordering, while a retrieval prior contributes task-skill semantic evidence that the contextual channel is not specialized to express.

\paragraph{Task-skill relevance prior.}
For each library skill $s_i$, let $
r(x,s_i) \;\in\; \mathbb{R} $
denote a task--skill relevance score produced by an off-the-shelf retriever $r$ applied to the task text as query and the skill metadata as documents.
We instantiate $r$ as TF--IDF cosine similarity over a unigram--bigram vocabulary built from the library; an ablation comparing this choice to BM25 and dense Qwen3-Embedding cosine is reported in \S\ref{sec:ablations}.
The scores $\{r(x,s_i)\}_{i=1}^K$ are precomputed once per task and reused at every decoding step at negligible cost, since each output index corresponds to fixed skill metadata.

\paragraph{Logit fusion at decoding time.}
At each decoding step $t$, the raw decoder logit $\ell_t(i)$ for skill index $i\in\{1,\dots,K\}$ is replaced by the fused logit, where $\bar r_i$ is a normalized or calibrated retrieval score.
\begin{equation}
\underbrace{\tilde{\ell}_t(i)}_{\text{fused logit}} \;=\; \underbrace{\ell_t(i)}_{\text{contextual}} \;+\; \alpha\cdot \underbrace{\bar r_i}_{\text{relevance}} \;+\; \beta\cdot \underbrace{\sigma_i}_{\text{set}},
\qquad i\in\{1,\dots,K\},
\label{eq:logit_fusion}
\end{equation}

with fixed scalars $\alpha,\beta\ge 0$ tuned on validation Set F1 (we use $\alpha{=}1.0$, $\beta{=}0.5$);
sensitivity to these weights is reported in \S\ref{sec:ablations}, where the surface is bowl-shaped within $\pm 2$\,pp Set F1 of the operating point.
We do not add retrieval or membership priors directly to the \texttt{STOP} logit. Termination is therefore controlled primarily by the AR stop logit and, when used, the cardinality prior. The fused logits $\tilde{\ell}_t$ are then passed to softmax and beam search as usual.

In sum, \ours yields three interpretable signals at every decoding step: a contextual term $\ell_t(i)$ that depends on the prefix $\hat{z}_{<t}$, a position-independent semantic relevance score $r(x,c,s_i)$, and a learned task-aware membership prior $\sigma_i$ that is order-agnostic.

\section{Experiment}
\label{sec:exp}

\subsection{Implementation Details}\label{sec:data}

\paragraph{Data curation.}
We assemble 9{,}872 task--skill-sequence training records over the curated skill library released with SkillBench~\cite{li2026skillsbench}, organised into three groups by ordering ground.
\textbf{Real anchors} are the 65 human-authored software-engineering tasks from SkillBench paired with gold skill annotations; per-task ordering is recovered from agent trajectory logs when available and from a Gemini~2.5~Pro fallback otherwise.
\textbf{Single-skill synthetic} tasks (2{,}880 records, synthesized by Gemini~2.5~Flash) cover all 196 skills uniformly and calibrate the composer to terminate after a single skill on simple queries.
\textbf{Multi-skill synthetic} tasks (6{,}927 records, synthesized by Gemini~2.5~Pro) cover compositions of 2--5 skills with two complementary ordering grounds: \textit{dependency} edges, where the upstream skill's output type overlaps the downstream skill's input type, encode hard data-flow ordering, and \textit{workflow} edges, mined from skill co-occurrence in real anchor trajectories, encode empirical ordering where no shared I/O type exists.
Both edge types are sampled from a 196-node skill dependency graph.
The full corpus is split 90\% into train set, and 5\% each for validation and test sets.
Synthesis prompts, the skill dependency graph, and the deduplication and validity-check pipeline are deferred to Appendix~\ref{app:data}.

\paragraph{Model implementation.}
As introduced in \S\ref{sec:method}, \ours pairs a frozen encoder with a small task-conditioned decoder.
We compare two encoder backbones: a causal LM (\textit{Qwen3-0.6B-Base}, mean-pooled) and a retrieval-tuned dense embedding model (\textit{Qwen3-Embedding-0.6B}, last-token pooled with the model card's instruct prefix).
The decoder is a 3-layer pre-norm Transformer with hidden width 256, 4 attention heads, dropout 0.1, and cross-attention into the 196-row skill memory; the output vocabulary is the closed library plus \texttt{STOP}, \texttt{START}, and \texttt{PAD}.
Two logits fusion priors are trained jointly with the sequence loss at weights $\alpha=0.5$ and $\beta=0.25$.
Training uses AdamW (learning rate $1{\times}10^{-4}$, weight decay 0.01, batch size 64) for up to 100 epochs with patience-15 early stopping on validation Set F1. At inference, we run a width-4 beam search with length penalty 0.7 and a duplicate-skill constraint, fusing a TF-IDF retrieval prior and the set-head score into the per-step decoder logits over the library indices only.
The lexical and set-fusion weights, together with a per-split stop bias that absorbs the synthetic-vs-real cardinality skew, are picked by coordinate ascent on the validation split.

\subsection{Experiment Setup}

\paragraph{Skill Composition Evaluation.}
We evaluate \ours under two regimes that share data composition but differ in the held-out subset, isolating the in-domain ceiling from real-task transfer.
SkillBench supplies the 196-skill library paired with 65 real tasks, and the graph-grounded synthetic corpus, all paired with deterministic verifiers built on the Harbor evaluation framework~\cite{Harbor_Framework}.
The \textbf{in-distribution test} ($n{=}494$) measures the ceiling when train and test draw from the same generator.
The \textbf{real-task holdout} ($n{=}65$) removes every real task from train and val, trains \ours and baseline methods on synthetic-only data, and tests on the 65 held-out real tasks.

\paragraph{Baselines.}
We compare \ours against three families of baselines, all predicting an ordered list of $\leq 8$ skill names from the same closed library.
\textit{Retrieval baselines} include BM25, TF--IDF cosine, and Qwen3-Embedding-0.6B; for each, we report a val-tuned-$k$ variant (best-$k$) and an \textit{oracle-$k$} variant given the gold list length, isolating selection quality from cardinality prediction.
\textit{LLM-judge (Gemini-2.5-flash)} is a frontier-API baseline that scores all 196 skills in a single prompt populated with their names and metadata, returning the ordered shortlist directly; the model picks both \textit{which} skills and \textit{how many}.
\textit{SFT Qwen3-0.6B-Base} fine-tunes the full 600M backbone to generate the ordered skill sequence as text, with the 196 skill names added to the tokenizer as special tokens and greedy decoding stopping at the trained EOS class.

\paragraph{Metrics.}
We report five metrics for skill prediction quality, covering selection and ordering, and report cardinality calibration separately in Figure~\ref{fig:cardinality}.
\textit{Set F1} is the order-agnostic F1 between predicted and gold skill sets, capturing selection quality and cardinality calibration in a single number.
\textit{Recall@5} measures gold-skill coverage in the top-5 predictions, decoupling selection from cardinality.
\textit{MRR} is the reciprocal rank of the first prediction that hits a gold skill.
\textit{nDCG@5} grades top-5 ordering by binary relevance with logarithmic discount.
\textit{Set EM} is order-agnostic exact match.
For downstream task performance, we report pass rate, normalised gain, and Codex input prompt tokens (Section~\ref{sec:exp}).

\paragraph{Task Performance Evaluation.}
To verify that better skill prediction translates to agent execution gains, we evaluate downstream task performance on 75 of the 88 SkillsBench tasks~\cite{li2026skillsbench}, excluding 13 office and document-processing tasks dominated by the Anthropic-bundled format skills (\texttt{pdf}, \texttt{xlsx}, \texttt{pptx}, \texttt{docx}) that any retriever trivially routes by file extension and that would therefore not discriminate between skill-loading methods.
We evaluate with two production-grade coding agents: \textbf{GPT-5.2-Codex} (via Azure OpenAI) and \textbf{Gemini-3-Pro-Preview} (via the Gemini CLI).
Each agent runs inside the Harbor evaluation framework~\cite{Harbor_Framework} with deterministic pytest verifiers and a 1200\,s timeout; we run three attempts per task (225 trials per agent--condition) at temperature 0 and report \textit{binary pass rate} following the SkillsBench protocol. We also report the average \textit{input prompt tokens} per non-errored trial as a measure of context overhead.
We compare \ours to four skill conditions: \textit{No~Skills} (no procedural context), \textit{All~Skills} (the full 196-skill library injected into the prompt), \textit{Retrieval~(top-3)} (from Qwen3-Embedding), and \textit{Retrieval~(oracle)} (retrieval restricted to the gold skill set), with \textit{Gold~Skills} (the curated task-specific oracle) as the upper bound.
\subsection{Skill Prediction Quality}

\begin{table*}[t]
\centering
\caption{\small Skill prediction quality (\%).
Left: in-distribution synthetic test ($n{=}494$). Right: real-task holdout ($n{=}65$); trained models are retrained on the real-task-removed partition.
Best non-oracle result in \textbf{bold}; second best \underline{underlined}; oracle-cardinality retrievers (in \textit{italics}) are reported as ceilings and excluded from the ranking.}
\label{tab:skill-prediction}
\setlength{\tabcolsep}{4pt}
\resizebox{\textwidth}{!}{\begin{tabular}{l ccccc c ccccc}
\toprule
& \multicolumn{5}{c}{\textbf{Synthetic test}}
& & \multicolumn{5}{c}{\textbf{Real-task holdout}} \\
\cmidrule(lr){2-6} \cmidrule(lr){8-12}
\textbf{Method}
& Set F1 & R@5 & MRR & nDCG@5 & SetEM
& & Set F1 & R@5 & MRR & nDCG@5 & SetEM \\
\midrule
\multicolumn{12}{l}{\textbf{\textit{Retrieval (oracle-$k$)}}}\\
\rowcolor{blue!15}
\textit{BM25}      & \textit{35.3} & \textit{35.3} & \textit{54.4} & \textit{38.3} & \textit{14.8} && \textit{55.9} & \textit{55.6} & \textit{73.6} & \textit{59.2} & \textit{33.8} \\
\rowcolor{blue!15}
\textit{TF--IDF}   & \textit{58.4} & \textit{58.4} & \textit{81.0} & \textit{63.0} & \textit{28.7} && \textit{74.2} & \textit{73.2} & \textit{89.2} & \textit{77.3} & \textit{47.7} \\
\rowcolor{blue!15}
\textit{Qwen3-Emb.} & \textit{49.0} & \textit{49.0} & \textit{69.9} & \textit{52.0} & \textit{20.2} && \textit{73.4} & \textit{72.7} & \textit{90.9} & \textit{76.9} & \textit{47.7} \\
\midrule
\multicolumn{12}{l}{\textbf{\textit{Retrieval (best-$k$)}}}\\
BM25 ($k{=}2$)             & 33.0 & 33.7 & 53.9 & 37.0 & \phantom{0}2.2 && 47.0 & 48.7 & 72.3 & 54.0 & \phantom{0}7.7 \\
TF--IDF ($k{=}2$)          & 52.5 & 53.0 & 82.2 & 58.9 & \phantom{0}4.3 && \underline{60.6} & 60.6 & 89.2 & \underline{67.7} & 10.8 \\
Qwen3-Emb. ($k{=}3$)        & 43.9 & 55.0 & 72.3 & 55.3 & \phantom{0}2.6 && 58.5 & \textbf{69.2} & \underline{90.8} & \textbf{73.8} & 10.8 \\
\midrule
\multicolumn{12}{l}{\textbf{\textit{LLM-judge}}}\\
Gemini-2.5-flash   & 61.0 & 69.0 & 69.3 & 63.1 & 21.3 && 59.9 & \underline{63.8} & 81.8 & 65.1 & 15.4 \\
\midrule
\multicolumn{12}{l}{\textbf{\textit{Trained models}}}\\
Qwen3-0.6B-Base  & \underline{71.1} & 68.9 & 79.2 & \underline{74.1} & \textbf{44.9} && 43.6 & 36.1 & 66.2 & 46.0 & \underline{16.9} \\
\rowcolor{gray!15}
\texttt{SkillComposer}$_{Base}$                & 70.4 & \underline{70.9} & \underline{84.2} & 73.2 & 37.2 && 53.9 & 45.0 & 87.7 & 54.1 & \underline{16.9} \\
\rowcolor{gray!15}
\texttt{SkillComposer}  & \textbf{73.9} & \textbf{72.4} & \textbf{86.5} & \textbf{75.0} & \underline{41.3}   && \textbf{62.9} & 54.7 & \textbf{90.8} & 63.4 & \textbf{20.0} \\
\bottomrule
\end{tabular}}
\end{table*}

\begin{wrapfigure}{r}{0.5\linewidth}
\vspace{-0.8em}
\centering
\includegraphics[width=\linewidth]{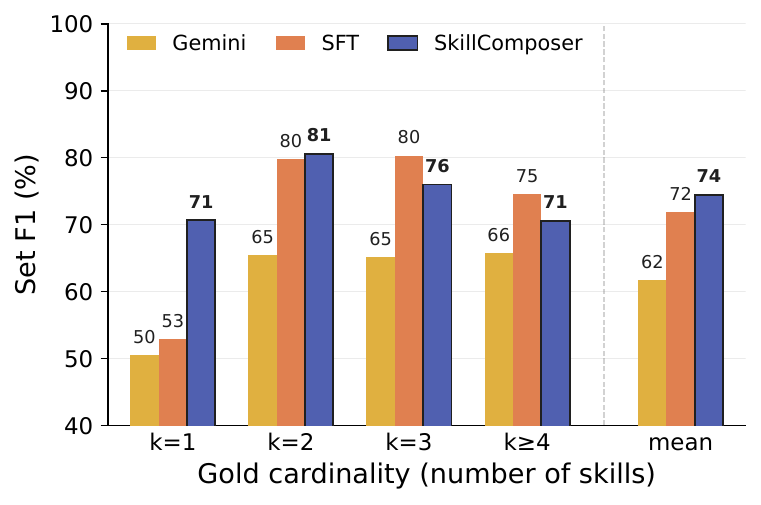}
\vspace{-6mm}
\caption{\small Cardinality robustness on the in-distribution test split. Set F1 stratified by gold cardinality $k$, with the macro-averaged mean on the right.}
\label{fig:cardinality}
\vspace{-0.8em}
\end{wrapfigure}

\paragraph{\ours wins in distribution at a fraction of the parameter cost.}
On the synthetic test (Table~\ref{tab:skill-prediction}, left), \ours leads SFT by $+2.8$\,pp Set F1 and the LLM-judge by $+12.9$\,pp while training $\sim$154$\times$ fewer parameters (3.9M vs 600M).
The same ranking holds on MRR and nDCG@5 — the AR head produces sharper top-of-list orderings even when SFT can directly memorise gold sequences as text, while SFT retains a small edge only on Set EM.
The cardinality slice (Fig.~\ref{fig:cardinality}) further shows that \ours's advantage concentrates in the $k{=}1$ bucket where over-emission is most punished, yielding the highest macro-averaged Set F1.
Both trained models dominate retrieval and the LLM-judge, including the oracle-$k$ retrieval ceilings, because they predict who-and-how-many jointly while retrieval needs the gold count and the LLM-judge cannot read full skill bodies within its context budget.

\paragraph{Under distribution shift, SFT degrades sharply while \ours degrades gracefully.}
On the real-task holdout (Table~\ref{tab:skill-prediction}, right), SFT loses 27.5\,pp going from synthetic to real tasks, whereas \ours loses only 11\,pp, a \textbf{$+19.3$\,pp Set F1 gap} on identical training data and library.
Among predicted-$k$ methods, only \ours approaches the oracle-$k$ ceilings without being told the gold count, and it remains the strongest on Set F1 even against the frontier LLM-judge.
Retrieval baselines actually improve from synthetic to real tasks because real-task phrasing is closer to the skill descriptions, whereas SFT has memorised the synthetic template distribution and has no robust prior to fall back on.
The frozen retrieval-tuned encoder paired with a small specialist decoder is what supplies \ours with this transfer bias.

\subsection{Downstream Task Performance}

\begin{wraptable}{r}{0.5\textwidth}
\centering
\vspace{-5mm}
\caption{\small Downstream task performance on SkillsBench.
Pass rate follows the paper-binary protocol, \textit{Tok.} is the average input prompt tokens per non-errored trial.
Best non-oracle result in \textbf{bold}; second best \underline{underlined}.}
\label{tab:downstream}
\setlength{\tabcolsep}{4pt}
\small
\resizebox{0.5\textwidth}{!}{
\begin{tabular}{l cc c cc}
\toprule
& \multicolumn{2}{c}{\textbf{GPT-5.2-Codex}}
& & \multicolumn{2}{c}{\textbf{Gemini-3-Pro}} \\
\cmidrule(lr){2-3} \cmidrule(lr){5-6}
\textbf{Skill condition}
& Pass\,(\%)\,$\uparrow$ & Tok.\,$\downarrow$
& & Pass\,(\%)\,$\uparrow$ & Tok.\,$\downarrow$ \\
\midrule
\rowcolor{blue!15}
\textit{Retrieval (oracle)}      & \textit{44.0} & \textit{1.13M} && \textit{42.2} & \textit{1.19M} \\
\rowcolor{blue!15}
\textit{Gold Skills}             & \textit{51.1} & \textit{1.12M} && \textit{48.4} & \textit{1.18M} \\
\midrule
No Skills                        & 22.2 & \textbf{0.94M} && 25.8 & \textbf{0.99M} \\
All Skills                       & 29.3 & 1.27M && 38.7 & 1.33M \\
Retrieval (top-$3$)              & \underline{44.0} & 1.09M && \underline{41.8} & 1.14M \\
\midrule
\rowcolor{gray!15}
\ours                            & \textbf{45.3} & \underline{1.03M} && \textbf{44.0} & \underline{1.08M} \\
\bottomrule
\end{tabular}}
\end{wraptable}

\paragraph{Skill prediction quality carries through to agent execution.}
Both agents follow the same pattern across baselines: pass rate climbs from \textit{No~Skills} to the \textit{Gold~Skills} ceiling, leaving roughly $+25$\,pp of headroom for any skill-loading mechanism to close.
Loading the entire library (\textit{All~Skills}) recovers only a small fraction of that headroom while inflating the Codex prompt to 1.27M input tokens, confirming that flooding the context is not enough.
\textit{Retrieval (top-$3$)} closes a much larger share at a smaller prompt budget, and even \textit{Retrieval (oracle)} only matches it, showing that the remaining headroom is driven by per-task selection quality rather than retrieval recall.
\ours predicts a calibrated ordered shortlist without using oracle skill labels, reaching $\mathbf{45.3}\,/\,\mathbf{44.0}$ pass rate on \{Codex, Gemini\}, beating both retrieval baselines and matching or exceeding \textit{Retrieval (oracle)}, while using the smallest prompt budget (1.03M Codex tokens) among skill-loaded conditions.
This closes roughly $80\%$ of the headroom on both agents, confirming that upstream gains in composition prediction (Table~\ref{tab:skill-prediction}) translate to real agent execution.

\subsection{Ablation Study}\label{sec:ablations}

All ablations train and evaluate on the in-distribution split; further ablations and per-layer breakdowns are reported in Appendix~\ref{app:ablations}.

\begin{figure}[t]
\centering
\begin{minipage}[b]{0.33\textwidth}
\centering
\includegraphics[width=\linewidth]{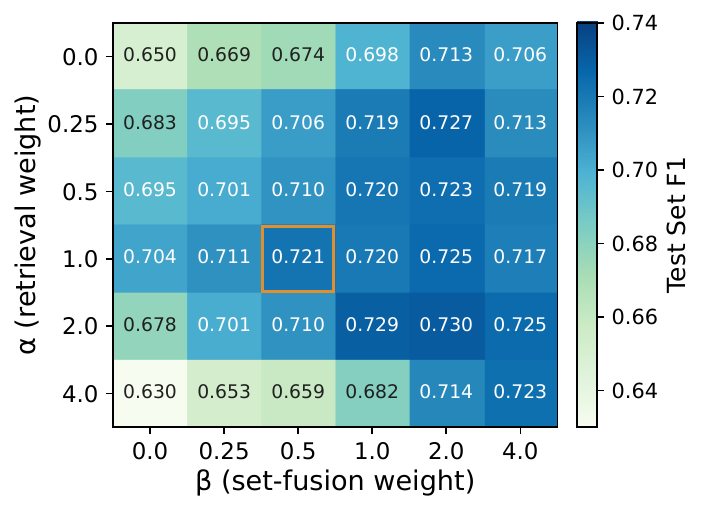}\\
{\small\textbf{(a)} $\alpha,\beta$ sensitivity grid.}
\end{minipage}\hfill
\begin{minipage}[b]{0.33\textwidth}
\centering
\includegraphics[width=\linewidth]{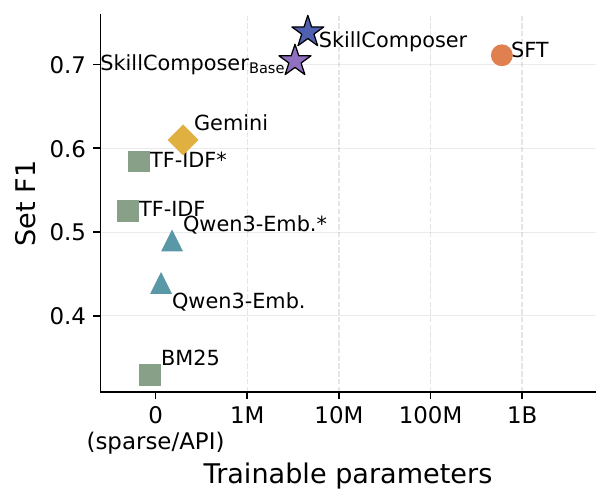}\\
{\small\textbf{(b)} Trainable params vs.\ accuracy.}
\end{minipage}\hfill
\begin{minipage}[b]{0.33\textwidth}
\centering
\includegraphics[width=\linewidth]{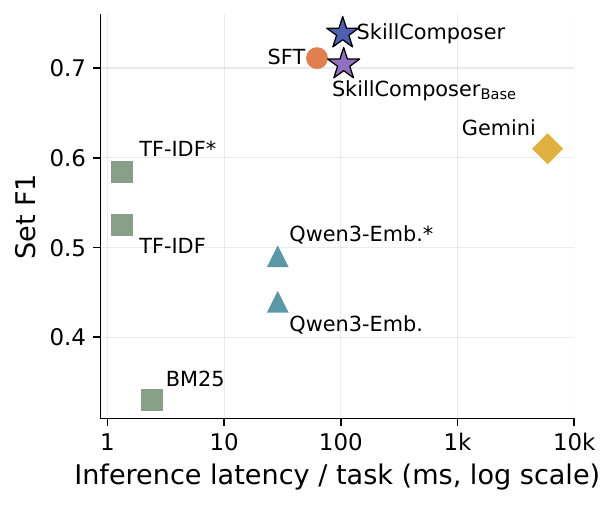}\\
{\small\textbf{(c)} Inference latency vs.\ accuracy.}
\end{minipage}
\caption{\small \textbf{(a)} Test Set F1 across the $(\alpha,\beta)$ decoding-weight grid; the surface is smooth and bowl-shaped around the val-selected operating point. \textbf{(b)} Compute--accuracy frontier on the synthetic test split: \ours w/ Qwen3-Embedding is Pareto-optimal among predicted-$k$ methods, sitting above SFT at $\sim$154$\times$ fewer trainable parameters and $\sim$25$\times$ less training compute. \textbf{(c)} Latency--accuracy frontier (1$\times$A6000, fp16, batch~1): \ours sits in the same latency class as SFT but with higher accuracy.}
\label{fig:ablate-figs}
\end{figure}

\paragraph{Hyperparameter sensitivity and compute--accuracy frontier.}
\begin{wraptable}{r}{0.35\textwidth}
\vspace{-1.2em}
\centering
\caption{\small Model component ablation.}
\label{tab:ablate-components}
\setlength{\tabcolsep}{4pt}
\small

\resizebox{0.35\textwidth}{!}{\begin{tabular}{l c}
\toprule
\textbf{Variant} & Set F1 \\
\midrule
AR-only (no auxiliary heads)              & 69.3 \\
\;\;{}+ set head                & 71.8 \\
\;\;{}+ cardinality head                & 69.6 \\
\midrule
\rowcolor{gray!15}
\ours                       & \textbf{73.9} \\
\;\;$-$ decode set-fusion ($\beta{=}0$)    & 65.0 \\
\;\;$-$ decode retrieval prior ($\alpha{=}0$) & 67.5 \\
\bottomrule
\end{tabular}}
\vspace{-1em}
\end{wraptable}
Figure~\ref{fig:ablate-figs}(a) sweeps decoding weights $(\alpha,\beta)$ on a $6{\times}6$ grid; the surface is smooth and bowl-shaped, with the val-selected operating point within $2$\,pp of every neighbour, indicating no fragile hand-tuning.
The compute--accuracy frontier (Fig.~\ref{fig:ablate-figs}b) shows \ours matching or beating SFT while training $\sim$154$\times$ fewer parameters and using $\sim$25$\times$ less compute, and dominating retrieval and the LLM-judge by a clear margin.
Inference latency (Fig.~\ref{fig:ablate-figs}c) sits in the same class as SFT and is two orders of magnitude faster than the API-based judge.
Together these views position \ours as Pareto-optimal among predicted-$k$ methods on this benchmark.

\paragraph{Each component is load-bearing.}
Table~\ref{tab:ablate-components} ablates components in \ours.
Starting from the AR head alone, adding the set-membership head during training lifts Set F1 by $+2.5$\,pp, order-agnostic gradients on every gold skill complement the AR objective.
At decode time, both fusion priors are necessary: zeroing the set-fusion bias costs $7.1$\,pp and zeroing the lexical retrieval prior costs $4.6$\,pp.
The auxiliary heads are reused at inference to refine the AR logits, and removing either signal degrades the predicted shortlist.

\begin{wraptable}{r}{0.4\textwidth}
\vspace{-1.7em}
\centering
\caption{\small Decode-time retrieval prior ablation.}
\label{tab:ablate-prior}
\setlength{\tabcolsep}{4pt}
\small
\begin{tabular}{l c}
\toprule
\textbf{Decode prior} & Set F1 \\
\midrule
No prior                          & 67.5 \\
BM25                          & 70.0 \\
Qwen3-Embedding       & 68.8 \\
\midrule
\rowcolor{gray!15}
\textbf{TF--IDF}              & \textbf{73.9} \\
\bottomrule
\end{tabular}
\vspace{-1em}
\end{wraptable}

\paragraph{Sparse beats dense as the decode-time prior.}
Table~\ref{tab:ablate-prior} sweeps the retrieval prior fed into the per-step decoder logits.
TF--IDF wins on Set F1 by $+2.5$\,pp over dense Qwen3-Embedding cosine and $+3.8$\,pp over no prior.
The closed library exposes 196 short, syntactically specific skill names; token-level overlap is high-precision for distinguishing them, while dense embeddings average over broader semantic context and overgeneralise.
The dense encoder is still the right choice for the task representation: \ours uses Qwen3-Embedding for $h_x$ and TF--IDF as the decode-time prior, combining the strengths of both.

\section{Conclusion}
We formalize skill composition as task-conditioned ordered skill-sequence prediction over a closed library, jointly resolving \textit{which} skills to load, \textit{how many}, and in \textit{what order}. \ours consists of a frozen retrieval-tuned encoder and a small autoregressive decoder whose logits fuse a TF-IDF retrieval prior and a set-membership signal at inference time. With only $\sim$3.9M trainable parameters, \ours matches SFT on an in-distribution test split and outperforms it by $+19.3$\,pp Set F1 on a held-out set of real software-engineering tasks, while remaining the strongest predicted-$k$ method against retrieval and frontier-API judges. The result suggests that, for closed agentic skill libraries, a small specialist that exploits the structure of the library is a more reliable composer than scaling up a generalist LM.

\section{Acknowledgement}
This project is supported by the Honda Research Institute USA.

\newpage
\bibliography{999_reference}
\bibliographystyle{style/icml2025}

\titlespacing*{\section}{0pt}{*1}{*1}
\titlespacing*{\subsection}{0pt}{*1.25}{*1.25}
\titlespacing*{\subsubsection}{0pt}{*1.5}{*1.5}

\setlength{\abovedisplayskip}{\baselineskip} 
\setlength{\abovedisplayshortskip}{0.5\baselineskip} 
\setlength{\belowdisplayskip}{\baselineskip}
\setlength{\belowdisplayshortskip}{0.5\baselineskip}

\clearpage
\appendix
\label{sec:append}
\part*{Appendix}
{
\setlength{\parskip}{-0em}
\startcontents[sections]
\printcontents[sections]{ }{1}{}
}

\setlength{\parskip}{.5em}
\section{Limitations and Broader Impacts}
\label{app:limit}

While \ours demonstrates that structured prediction over a reusable skill library yields accurate and well-calibrated skill compositions across diverse agent task domains, our study mainly focuses on text-only task descriptions paired with a code-oriented skill library, evaluated through composition-level metrics and downstream agent benchmarks. A more comprehensive exploration of structured skill composition could incorporate multimodal task specifications (screenshots, sketches, voice instructions), interactive and long-horizon settings where the skill library is updated online, and specialized domains such as scientific workflows, robotics, and embodied agents where the composition graph extends across heterogeneous tools and physical actuators. Another direction left to future work is scaling the underlying composer beyond the small-LM and embedding backbones used here; \ours inherits the characteristics of these components, including the linguistic priors and task coverage of the pre-training corpora, and we expect stronger backbones and larger curated skill libraries to further sharpen composition accuracy and ordering.

Regarding the broader impacts of \ours, it advances flexible and efficient agent construction by predicting which reusable skills to compose, how many, and in what dependency order, making it well-suited for real-world scenarios where developers maintain growing skill libraries and need reliable orchestration, such as data analysis assistants, web-task automation, and database interaction. Operating at the skill abstraction (rather than raw API calls) also encourages reuse and modular auditing, reducing redundant code generation and promoting more sustainable deployment of agent systems. We follow standard practices in model development and evaluation, use only public datasets and openly released model checkpoints, and encourage responsible use in downstream applications. This work does not target any specific sensitive domain and is intended as a general-purpose framework for advancing structured skill composition for LLM agents.

\section{Implementation Details}
\label{app:data}

This appendix expands Section~\ref{sec:data} with the skill dependency graph, the Gemini prompts used for synthesis, and the deduplication and validation pipeline.

\subsection{Skill Dependency Graph}
\label{app:graph}

The graph that grounds multi-skill synthesis has 196 nodes (one per library skill) and two structural edge types used for sampling, summarised in Table~\ref{tab:skill-graph}.
\textit{Dependency} edges connect skill pairs whose input/output schemas overlap: the upstream skill's output type matches an input type of the downstream skill, so the ordering is determined by data flow.
\textit{Workflow} edges connect skill pairs that co-occur in real-task agent trajectories; ordering follows the observed execution order, used when no shared I/O type exists.
Sampling for multi-skill synthesis draws 65\,\% of pairs from dependency edges and 35\,\% from workflow edges, matching the observed ratio of hard data-flow chains to looser plan-then-implement workflows in real anchor tasks.

\begin{table}[h]
\centering
\caption{\small Skill dependency graph used for grounding multi-skill synthesis.}
\label{tab:skill-graph}
\begin{tabular}{lr}
\toprule
\textbf{Edge type} & \textbf{Count} \\
\midrule
Dependency (I/O overlap) & 658 \\
Workflow (anchor co-occurrence) & 266 \\
\midrule
Total & 924 \\
\bottomrule
\end{tabular}
\end{table}

\subsection{Synthesis Prompts}
\label{app:prompts}

We synthesise single-skill records (Figure~\ref{fig:prompt-l2}) and multi-skill records (Figure~\ref{fig:prompt-l3}) with two Gemini prompts.
The single-skill prompt asks Gemini~2.5~Flash to produce five tasks per call across distinct domains and difficulty levels, never naming the target skill in the task description, so the composer has to recover skill identity from semantics rather than surface form.
The multi-skill prompt asks Gemini~2.5~Pro to compose all input skills into one realistic task, propose an execution order, and emit a permutation of the supplied skill IDs; whenever sampled skills are connected by a dependency edge, an explicit ordering constraint is injected so that the data-flow direction is preserved.

\begin{figure*}[hp]
\centering
\caption{Single-skill synthesis prompt (Gemini~2.5~Flash). Five tasks per call ensure scenario diversity at fixed cost; difficulty is balanced 2/2/1 across easy/medium/hard.}
\label{fig:prompt-l2}
\begin{tcolorbox}[colback=lightgray!10, colframe=black, width=\textwidth, arc=2mm, boxrule=0.5mm, title=Single-skill synthesis]
\begin{lstlisting}[basicstyle=\small\ttfamily, breaklines=true]
You write realistic task descriptions for an AI coding agent.

Given this skill:
  Name: {name}
  Description: {description}
  Body (truncated): {body_truncated}

Generate EXACTLY 5 realistic task descriptions that a user would give to an
AI agent, where solving each task requires this exact skill. Requirements:
- Do NOT name the skill directly in any task description.
- Each task must use a DIFFERENT scenario/domain (e.g., finance, IoT
  sensors, bioinformatics, logistics, social media analytics).
- Each task must use a DIFFERENT input shape (e.g., single file, directory
  of files, streaming data, API response, database export).
- No two tasks should share the same core use case or workflow pattern.
- Each task should be concrete enough that a solution can be evaluated.
- Prefer domain-realistic framing (a user's workflow, a data scenario, ...).
- Difficulty distribution: exactly 2 easy, 2 medium, and 1 hard task.

Return STRICT JSON only, no markdown fences, no extra text -- a JSON array
of exactly 5 objects:
[
  {"task": "...", "difficulty": "easy|medium|hard",
   "reasoning": "why this skill is needed",
   "scenario_tag": "short domain label"},
  ...
]
\end{lstlisting}
\end{tcolorbox}
\end{figure*}

\begin{figure*}[hp]
\centering
\caption{Multi-skill synthesis prompt (Gemini~2.5~Pro). Ordering constraints from dependency edges are injected verbatim when present; otherwise, Gemini proposes an execution order with rationale.}
\label{fig:prompt-l3}
\begin{tcolorbox}[colback=lightgray!10, colframe=black, width=\textwidth, arc=2mm, boxrule=0.5mm, title=Multi-skill synthesis with dependency constraints]
\begin{lstlisting}[basicstyle=\small\ttfamily, breaklines=true]
You write realistic tasks for an AI coding agent that require composing
multiple skills.

Given these {k} skills:
{skill_list}

Generate a realistic task description that requires ALL {k} skills to
solve, composed in a specific execution order. Requirements:
- Do NOT name the skills directly in the task description.
- Specify which skill runs first, second, etc., with brief rationale.
- The task should be concrete and domain-realistic (data pipeline,
  analysis workflow, system integration, etc.).

ORDERING CONSTRAINTS (from dependency analysis):
- "<skill_A>" MUST come before "<skill_B>" (produces data that <skill_B>
  consumes)
... (one line per dependency edge among the sampled skills)
You MUST respect these ordering constraints when determining execution
order.

Return STRICT JSON only (no markdown fences, no extra text):
{
  "task": "<concrete task description>",
  "ordered_skills": [<all skill IDs, in chosen execution order>],
  "rationale": "<why this execution order>",
  "difficulty": "easy|medium|hard"
}

IMPORTANT: ordered_skills must be a permutation of the supplied IDs --
same elements, different order based on execution dependencies.
\end{lstlisting}
\end{tcolorbox}
\end{figure*}

\subsection{Deduplication and Validation}
\label{app:quality}

Every synthesised record is checked against all previously accepted records before being added to the pool, using three layers of similarity in escalating strictness.
First, an exact-string match on the task identifier or instruction text rejects byte-identical duplicates.
Second, character-trigram Jaccard similarity above 0.6 between two instructions is treated as near-duplicate phrasing and rejected.
Third, sentence-embedding cosine similarity above 0.92, computed against the cached library embedding bank, is treated as a semantic duplicate.
Records that survive deduplication are validated against the closed vocabulary: any response that adds, drops, or renames a skill — or whose \texttt{ordered\_skills} field is not an exact permutation of the prompt input — is dropped.
For multi-skill records, we additionally check that any dependency-edge ordering constraint emitted into the prompt is respected by the returned ordering.
The 90/5/5 train/val/test split is applied independently within each of the three groups (real anchors, single-skill synthetic, multi-skill synthetic) using a fixed seed (42), so the group ratio is preserved across splits and the validation/test sets are not dominated by the more numerous synthetic groups.

\section{Extra Analysis}
\label{sec:analysis}

\subsection{Downstream Case Studies}
\label{sec:case-studies}

To understand where the headline pass-rate gap in Table~\ref{tab:downstream} comes from, we inspect three SkillsBench tasks (GPT-5.2-Codex, three trials each, identical agent and task definition) where \ours, \textit{Retrieval (top-$3$)}, and \textit{Gold Skills} disagree on the supplied skill set.
The three cases isolate three distinct mechanisms behind the gap.

\paragraph{Case 1: top-$3$ truncation drops a key skill, and \ours diverges from gold.}
On \texttt{adaptive-cruise-control}, top-$3$ retrieval keeps the three obvious control skills but cuts \texttt{imc-tuning-rules} — the IMC heuristic that produces PID gains satisfying the rise-time and overshoot specs in one shot — so the agent hand-tunes gains and misses the spec on 2/3 trials.
The curated \textit{Gold Skills} set is even more telling: it bundles two I/O-format skills (\texttt{csv-processing}, \texttt{yaml-config}) that add no information for the controller-tuning bottleneck, and itself only reaches 0.33.
\ours diverges from gold, drops the I/O wrappers, and adds the substantive \texttt{imc-tuning-rules} that gold itself failed to include, recovering full reward.
The case shows that \ours is not regressing to the gold key but identifying useful skills, while top-$k$ retrieval, by construction, can always be one slot short of the skill that turns a near-miss into a pass.

\begin{tcolorbox}[colback=lightgray!10, colframe=black, width=\textwidth, arc=2mm, boxrule=0.5mm, title=\textbf{Example 1.} \texttt{adaptive-cruise-control}]
\small
\label{exa:case1}
\textit{Task.} Implement an Adaptive Cruise Control simulation; the verifier checks rise-time $<\!10\,\mathrm{s}$, overshoot $<\!5\%$, steady-state speed error $<\!0.5\,\mathrm{m/s}$, distance steady-state error $<\!2\,\mathrm{m}$, and minimum gap $>\!5\,\mathrm{m}$.
\vspace{2pt}
\begin{center}
\setlength{\tabcolsep}{4pt}
\renewcommand{\arraystretch}{1.15}
\begin{tabular}{l c >{\raggedright\arraybackslash}p{0.62\linewidth}}
\toprule
\textbf{System} & Reward & Skills supplied to agent \\
\midrule
\rowcolor{gray!15}
\ours              & \textbf{1.00} & \texttt{imc-tuning-rules}, \texttt{pid-controller}, \newline \texttt{simulation-metrics}, \texttt{vehicle-dynamics} \\
Retrieval (top-$3$) & 0.33          & \texttt{pid-controller}, \texttt{simulation-metrics}, \texttt{vehicle-dynamics} \\
Gold Skills        & 0.33          & \texttt{csv-processing}, \texttt{pid-controller}, \texttt{simulation-metrics}, \newline \texttt{vehicle-dynamics}, \texttt{yaml-config} \\
\bottomrule
\end{tabular}
\end{center}
\end{tcolorbox}

\paragraph{Case 2: a leaner skill set beats gold.}
On \texttt{exoplanet-detection-period}, \ours converges on the minimal preprocess\,$\to$\,Lomb-Scargle\,$\to$\,TLS recipe and solves the task on every trial, while the gold pack adds the redundant \texttt{box-least-squares} estimator and the heavyweight \texttt{exoplanet-workflows} wrapper, which lead the agent down a longer pipeline that overfits the stellar oscillation.
This confirms that \ours is learning which skills are actually useful, and that smaller well-chosen sets can beat larger curated ones — consistent with the \textit{All Skills} row in Table~\ref{tab:downstream}, where dumping the full library hurts pass rate.

\begin{tcolorbox}[colback=lightgray!10, colframe=black, width=\textwidth, arc=2mm, boxrule=0.5mm, title=\textbf{Example 2.} \texttt{exoplanet-detection-period}]
\small
\label{exa:case2}
\textit{Task.} A TESS lightcurve hides an exoplanet signal under stellar-activity oscillations; recover the planet's orbital period.
\vspace{2pt}
\begin{center}
\setlength{\tabcolsep}{4pt}
\renewcommand{\arraystretch}{1.15}
\begin{tabular}{l c >{\raggedright\arraybackslash}p{0.62\linewidth}}
\toprule
\textbf{System} & Reward & Skills supplied to agent \\
\midrule
\rowcolor{gray!15}
\ours       & \textbf{1.00} & \texttt{light-curve-preprocessing}, \newline \texttt{lomb-scargle-periodogram}, \texttt{transit-least-squares} \\
Gold Skills & 0.00          & \texttt{box-least-squares}, \texttt{exoplanet-workflows}, \newline + the three above \\
\bottomrule
\end{tabular}
\end{center}
\end{tcolorbox}

\paragraph{Case 3: short-sequence bias under-emits on long-chain tasks.}
On \texttt{lean4-proof}, \ours emits a single Lean-related skill (\texttt{lean4-memories}, the snippet-level memo skill) and stops before \texttt{lean4-theorem-proving}, the tactic and Mathlib reference skill that the agent needs to discharge the inductive step.
Retrieval keeps the full three-skill chain (adding \texttt{python-scala-functional}) and reaches 1.00; gold has the two-skill chain and matches \ours at 0.67 only because of an unrelated Lean kernel error on one trial.
The same one-slot-short pattern recurs on \texttt{grid-dispatch-operator} and \texttt{dapt-intrusion-detection}: whenever the gold sequence has $\geq\!2$--$3$ skills, \ours's predicted shortlist tends to fall a slot short, reflecting the synthetic corpus's emphasis on $\leq\!3$-skill compositions during data construction.
This points to the most actionable headroom — improving the construction of long-sequence training records — and is consistent with the per-cardinality breakdown~\ref{fig:cardinality}, where SFT (with full LM-style decoding) holds a small edge on $k{\geq}3$ buckets.

\begin{tcolorbox}[colback=lightgray!10, colframe=black, width=\textwidth, arc=2mm, boxrule=0.5mm, title=\textbf{Example 3.} \texttt{lean4-proof}]
\small
\label{exa:case3}
\textit{Task.} Finalise a Lean~4 proof template that $S_n=\sum_{i=0}^{n} 1/2^i \le 2$ for all $n\in\mathbb{N}$.
\vspace{2pt}
\begin{center}
\setlength{\tabcolsep}{4pt}
\renewcommand{\arraystretch}{1.15}
\begin{tabular}{l c >{\raggedright\arraybackslash}p{0.6\linewidth}}
\toprule
\textbf{System} & Reward & Skills supplied to agent ($|\cdot|$) \\
\midrule
\rowcolor{gray!15}
\ours              & 0.67          & \texttt{lean4-memories} \hfill (1) \\
Retrieval (top-$3$) & \textbf{1.00} & \texttt{lean4-memories}, \texttt{lean4-theorem-proving}, \newline \texttt{python-scala-functional} \hfill (3) \\
Gold Skills        & 0.67          & \texttt{lean4-memories}, \texttt{lean4-theorem-proving} \hfill (2) \\
\bottomrule
\end{tabular}
\end{center}
\end{tcolorbox}

\section{Additional Ablations}
\label{app:ablations}

This appendix expands Section~\ref{sec:ablations} with a per-layer view of the synthetic test split for the deterministic baselines.

\subsection{Per-layer breakdown of deterministic baselines}
\label{app:per-layer}

Table~\ref{tab:appendix-per-layer} reports Set F1 per data layer on the synthetic test split for the deterministic (retrieval and LLM-judge) baselines.
The synthetic test contains only $n{=}4$ real-anchor records; with such a small sample, any per-layer estimate has a standard error above $0.25$, so the real-anchor column should be read with caution.
On the synthetic-only layers, the LLM-judge wins on the multi-skill compositional layers (free-form 81.6, graph-grounded 75.0) where the prompt's longer task descriptions give it more semantic signal, while TF--IDF and Qwen3-Embedding retrieval cluster around 0.45--0.55.

\begin{table}[h]
\centering
\caption{\small Per-layer Set F1 on the synthetic test split (\%, deterministic baselines only).
Trained-model rows are omitted because the per-layer canonical predictions were not saved with matching record IDs; rerunning inference on the canonical checkpoints is left to a future revision.}
\label{tab:appendix-per-layer}
\setlength{\tabcolsep}{5pt}
\begin{tabular}{l cccc c}
\toprule
\textbf{Method} & Real anchors & Single-skill  & Multi-skill  & Multi-skill (graph)  & All \\
\midrule
TF--IDF ($k{=}2$)               & 53.3 & 45.8 & 65.9 & 53.7 & 52.5 \\
\textit{TF--IDF (oracle-$k$)}   & \textit{66.7} & \textit{58.3} & \textit{66.8} & \textit{57.1} & \textit{58.4} \\
BM25 ($k{=}2$)                  & 29.2 & 31.0 & 35.0 & 33.6 & 33.0 \\
Qwen3-Emb ($k{=}3$)             & 51.8 & 36.8 & 50.3 & 46.3 & 43.9 \\
\textit{Qwen3-Emb (oracle-$k$)} & \textit{62.5} & \textit{52.8} & \textit{50.4} & \textit{46.8} & \textit{49.0} \\
LLM-judge         & 62.5 & 60.5 & 81.6 & 75.0 & 71.3 \\
\bottomrule
\end{tabular}
\end{table}

\end{document}